# A note on 'A fully parallel 3D thinning algorithm and its applications'[1]

Tao Wang and Anup Basu

**Department of Computing Science, University of Alberta, Edmonton, AB T6G 2E8, Canada**
**Email: {taowang, anup}@cs.ualberta.ca**

**Abstract**

A 3D thinning algorithm erodes a 3D binary image layer by layer to extract the skeletons. This paper presents a correction to Ma and Sonka's thinning algorithm, 'A fully parallel 3D thinning algorithm and its applications', which fails to preserve connectivity of 3D objects. We start with Ma and Sonka's algorithm and examine its verification of connectivity preservation. Our analysis leads to a group of different deleting templates, which can preserve connectivity of 3D objects.

**1. Introduction**

*Thinning* is a useful technique having potential applications in a wide variety of problems. It creates a compact representation (*skeleton*) of the models that may be used for further processing. 3D skeletons can be used in many applications [1-2] such as 3D pattern matching, 3D recognition and 3D database retrieval.

A *3D binary image* is a mapping that assigns the value of 0 or 1 to each point in the 3D space. Points having the value of 1 are called *black (object)* points, while 0's are called *white (background)* ones. Black points form objects of the binary image. The thinning operation iteratively *deletes* or *removes* some object points (that is, changes some black points to white) until only some restrictions prevent further operation. Note that the white points will never be changed to black ones in any circumstances. Most of the existing thinning algorithms are parallel, since the *medial axis transform* (MAT) can be defined as *fire front propagation*, which is by nature parallel [3]. There are three categories of parallel thinning algorithms in literature, *sub-iteration parallel* thinning algorithm [4-6], *sub-field parallel* thinning algorithm [7-8] and *fully parallel* thinning algorithm [9]. Brief surveys of algorithms in each category can be found in the literature [6, 9].

The rest of this paper is organized as follows. In Section 2, some basic concepts will be presented. Section 3 will briefly discuss Ma and Sonka's algorithm [9]. The problematic part in this algorithm is analyzed and the modification is presented in Section 4, before the work is concluded in Section 5.

---

[1] The support of an ICORE scholarship in making this work possible is gratefully acknowledged.

## 2. Basic concepts

We first describe some terms and notation:

Let *p* and *q* be two different points with coordinates (*px*, *py*, *pz*) and (*qx*, *qy*, q*z*), respectively, in a 3D binary image *P*. The Euclidean distance between *p* and *q* is defined as:

$$dis = \sqrt{(px-qx)^2 + (py-qy)^2 + (pz-qz)^2}$$

Then *p* and *q* are:

*6-adjacent* if $dis \leq 1$. *18-adjacent* if $dis \leq \sqrt{2}$. *26-adjacent* if $dis \leq \sqrt{3}$. Let us denote the set of points *k-adjacent* to point *p* by $N_k(p)$, where *k* = 6, 18, 26, (see Figure 1). $N_k(p)$ is also called *p*'s *k-neighborhood*. Let *p* be a point in a 3D binary image. Then, *e(p)*, *w(p)*, *n(p)*, *s(p)*, *u(p)*, and *d(p)* are *6-neighbors* of *p*, which represent 6 directions of *east*, *west*, *north*, *south*, *up*, and *down*, respectively. The *18-neighbors* of *p* (but not in *p*'s *6-neighborhood*) are *nu(p)*, *nd(p)*, *ne(p)*, *nw(p)*, *su(p)*, *sd(p)*, *se(p)*, *sw(p)*, *wu(p)*, *wd(p)*, *eu(p)*, and *ed(p)*, which represent 12 directions of *north-up, north-down, north-east, north-west, south-up, south-down, south-east, south-west, west-up, west-down, east-up,* and *east-down*, respectively.

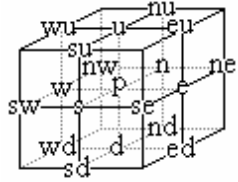

**Figure 1:** The adjacencies in a 3D binary image. Points in $N_6(p)$ are marked *u, n, e, s, w,* and *d*. Points in $N_{18}(p)$ but not in $N_6(p)$ are marked *nu, nd, ne, nw, su, sd, se, sw, wu, wd, eu,* and *ed*. The unmarked points are in $N_{26}(p)$ but not in $N_{18}(p)$.

It is very important for thinning algorithms to *preserve connectivity* for 3D objects [4, 9]. If a thinning algorithm fails to preserve connectivity, the skeletons extracted from the object will be disconnected, which is unacceptable in many applications. A sequential thinning algorithm can preserve connectivity easily if it is only allowed to delete *simple points* [10-12]. However, a parallel thinning algorithm may delete many black points in every iteration, even if it is only allowed to delete simple points, the algorithm may not preserve connectivity. This problem was investigated in [13-15].

## 3. Ma and Sonka's algorithm

In 1996, Ma and Sonka proposed a fully 3D thinning algorithm [9], which was applied to many applications such as medical image processing [16] and 3D reconstruction [17].

The algorithm is based on some pre-defined templates (Class A, B, C and D). If the neighborhood of an object point matches one of the templates, it will be removed. Figure 2 shows the four basic template cores. In this figure, a "•" is used to denote an object point, a "∘" is used to denote a background point. An unmarked point is a "don't care" point, which can represent either an object point or a background point.

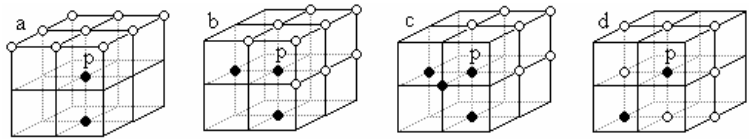

**Figure 2:** Four template cores (Class A, B, C and D) of the fully parallel thinning algorithm. A "•" is an object point. A "∘" is a background point. An unmarked point is a "don't care" point that can be either an object point or a background point. For (d), there is an additional restrict that $p$ must be a simple point.

The template cores themselves are not the deleting templates. Some translations [9] must be applied to the template cores to generate the deleting templates. There are 6 templates in Class A, 12 templates in Class B and 8 templates in Class C and 12 templates in Class D according to the translations. Templates in Class A are shown in [9], Templates in Class B-D are shown in Figure 4-Figure 6. In Figure 6, at least one point marked ? is an object point.

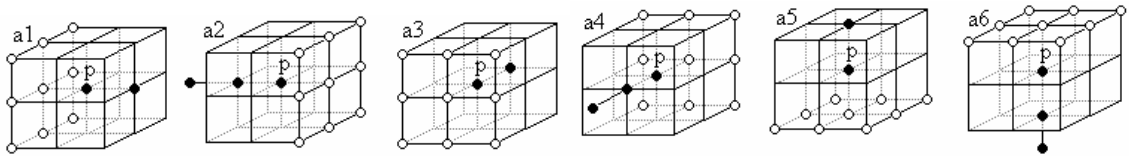

**Figure 3:** 6 deleting templates in Class A.

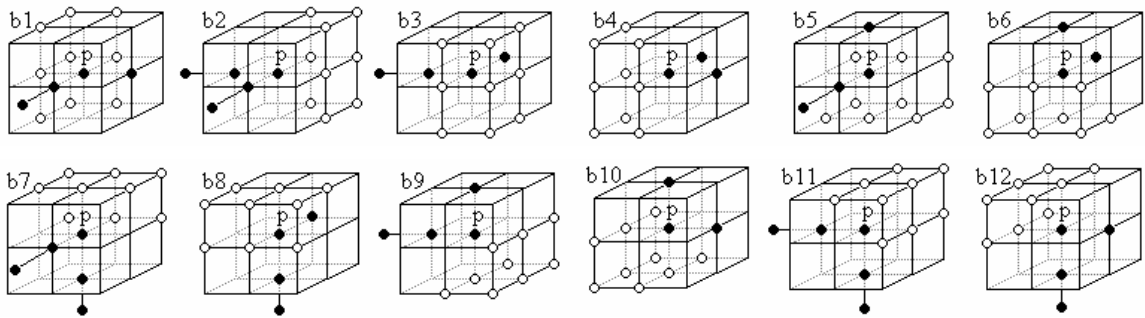

**Figure 4:** 12 deleting templates in Class B.

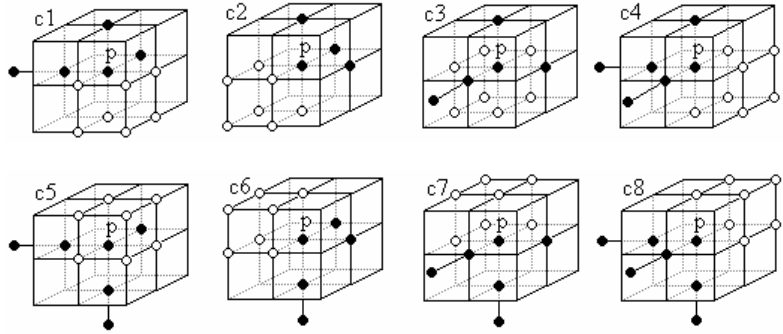

**Figure 5:** 8 deleting templates in Class C.

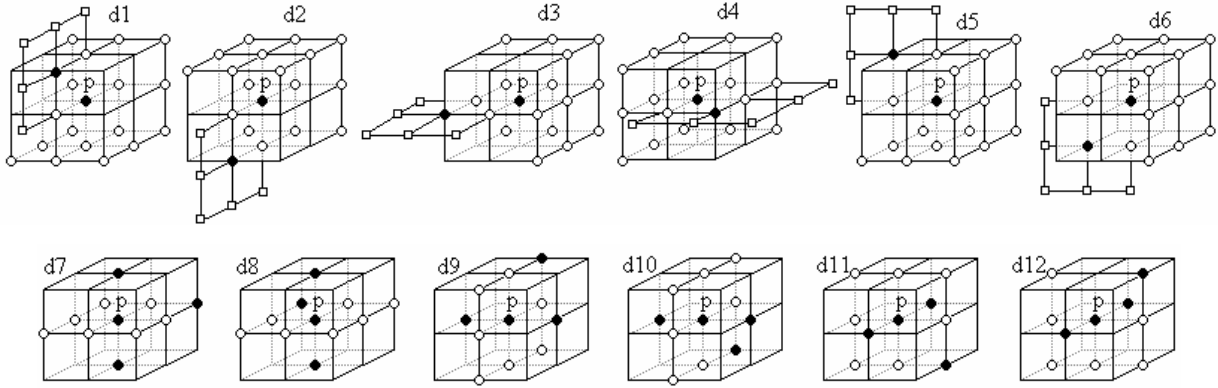

**Figure 6:** 12 deleting templates in Class D, where at least one point marked ? is an object point

Ma and Sonka defined the preserving conditions as follows:

"Rule 2.2. Let *p* be an object point of a 3D image. Then:

1. *p* is called a line-end point if p is 26-adjacent to exactly one object point,

2. *p* is called a near-line-end point if p is 26-adjacent to exactly two object points which are:

(a) either $s(p)$ and $e(p)$, or $s(p)$ and $u(p)$ but not both;

(b) either $n(p)$ and $w(p)$, or $u(p)$ and $w(p)$ but not both; or

(c) either $n(p)$ and $d(p)$, or $e(p)$ and $d(p)$ but not both;

3. p is called a *tail point* if it is either a line-end point near-line-end point; otherwise it is called a *nontail point*.", where $e(p), w(p), n(p), s(p), u(p)$ and $d(p)$ are the east, west, north, south, up, and down neighbors of *p*, respectively.

In each iteration, all non tail-points satisfying at least one of the deleting templates in Class A, B, C or D are deleted in the fully parallel thinning algorithm as follows:

**Algorithm**
**Repeat**
   1) Mark every object point which is 26-adjacent to a background point;
   2) **Repeat**
      *Simultaneously delete every non tail-point which satisfies at least one deleting template in Class A, B, C, or D;*
   **Until** no point can be deleted;
   3) Release all marked but not deleted points;
**Until** no marked point can be deleted;

## 4. The problem and a solution

A 3D parallel thinning algorithm should preserve connectivity. However, by studying the configuration in Figure 7, we find that the algorithm fails to preserve connectivity. In Figure 7, a "●" is an object point. All other points are background points, and it shows a 26-connected 3D object a-b-c-d-e-f-g. In Ma and Sonka's thinning algorithm, point $c$, $d$ and $e$ will be deleted because $c$ satisfies template $a5$ in Class A, $d$ satisfies template $d7$ in Class D and $e$ satisfies template $a6$ in Class A. However, the deletion of point $c$, $d$ and $e$ leads to disconnection of the object.

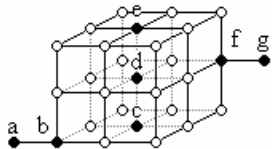

**Figure 7:** A connected object a-b-c-d-e-f-g in 3D space. A "●" is an object point. A "○" is a background point. All other points in 3D space are background points. In Ma and Sonka's algorithm, point $c$ will be deleted by template $a5$ in Class A, point $d$ will be deleted by template $d7$ in Class D and point $e$ will be deleted by template $a6$ in Class A. Hence, the object will be disconnected.

Lohou discovered this problem and gave a counter example of Ma and Sonka's algorithm in [18]. Some other researchers, such as Chaturvedi [16] who applied this algorithm and found it disconnected small segments, but did not know how to fix this problem. In this section, we will show the reason for the problem and how to modify the templates in Class D to preserve connectivity.

Ma and Sonka proposed a general theorem [9] and used it to prove that the 3D thinning algorithm preserves connectivity in the VERIFICATION section in that paper. According to our observation, we note that LEMMA 3.5 in the VERIFICATION section is problematic.

"**LEMMA 3.5**: *Let p, q be two 6-adjacent object points in a 3D binary image where both p and q satisfy*

$\Omega$. *Then either $q \notin \Omega(p)$ or $p \notin \Omega(q)$.*"

$\Omega$ is used to denote the set of deleting templates in Class A, B, C or D. An object point satisfies $\Omega$ if it satisfies any one of the deleting templates in $\Omega$. "$q \in \Omega(p)$" means that $q$ must be an object point for $p$ to satisfy $\Omega$. "$q \notin \Omega(p)$" means $p$ still satisfies $\Omega$ after $q$ is deleted.

For the 3D object in Figure 7, $c$ and $d$ are two 6-adjacent points. According to LEMMA 3.5, either $c \notin \Omega(d)$ or $d \notin \Omega(c)$. However, if $c$ is deleted, $d$ will not satisfy any of the deleting templates. And if $d$ is deleted, $c$ will not satisfy any of the deleting templates. Therefore, although $c$ and $d$ are 6-adjacent, $c \in \Omega(d)$ and $d \in \Omega(c)$. We can prove that for points $d$ and $e$, $d \in \Omega(e)$ and $e \in \Omega(d)$, although $d$ and $e$ are 6-adjacent, in the same way.

LEMMA 3.5 requires that for two 6-adjacent points $p$ and $q$, if both $p$ and $q$ satisfy $\Omega$, then either $q \notin \Omega(p)$ or $p \notin \Omega(q)$. Let *p1* and *p2* be the two "don't care" points in *p*'s 6-neighborhood, as showed in Figure 8. According to LEMMA 3.5, if *p1* is 1, then *p2* must be 0; if *p2* is 1 then *p1* must be 0. So (*p1, p2*) can be (0, 0), (0, 1) or (1, 0), but not (1, 1). There is no template in Class A-C that has value of (1, 1) for (*p1, p2*), however, the deleting templates in Class D violate this rule. For instance, in template *d7*, (*p1, p2*) is (1, 1), which causes LEMMA 3.5 to fail.

Based on this observation, we can change deleting template *d1-d12* to make LEMMA 3.5 satisfy. For instance, we change template *d7* to three new templates as shown in Figure 9, according to different values of (*p1, p2*).

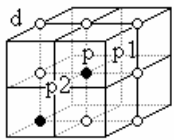  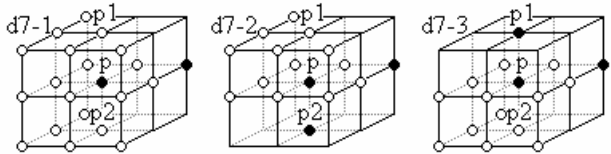

**Figure 8**: Template core of Class D.         **Figure 9**: Template d7-1 to d7-3.

In this way, we change the 12 deleting templates of Class D to 36 deleting templates according to different values of (*p1, p2*). So there are 36 templates in Class D, and 6 + 12 + 8 + 36 = 62 templates in total. Figure 10 shows the modified templates in Class D. Each template in Class D is changed to three templates, in which (*p1, p2*) are (0, 0), (0, 1) or (1, 0) respectively. Since (*p1, p2*) is not (1, 1) in any template, LEMMA 3.5 is satisfied for the new set of templates.

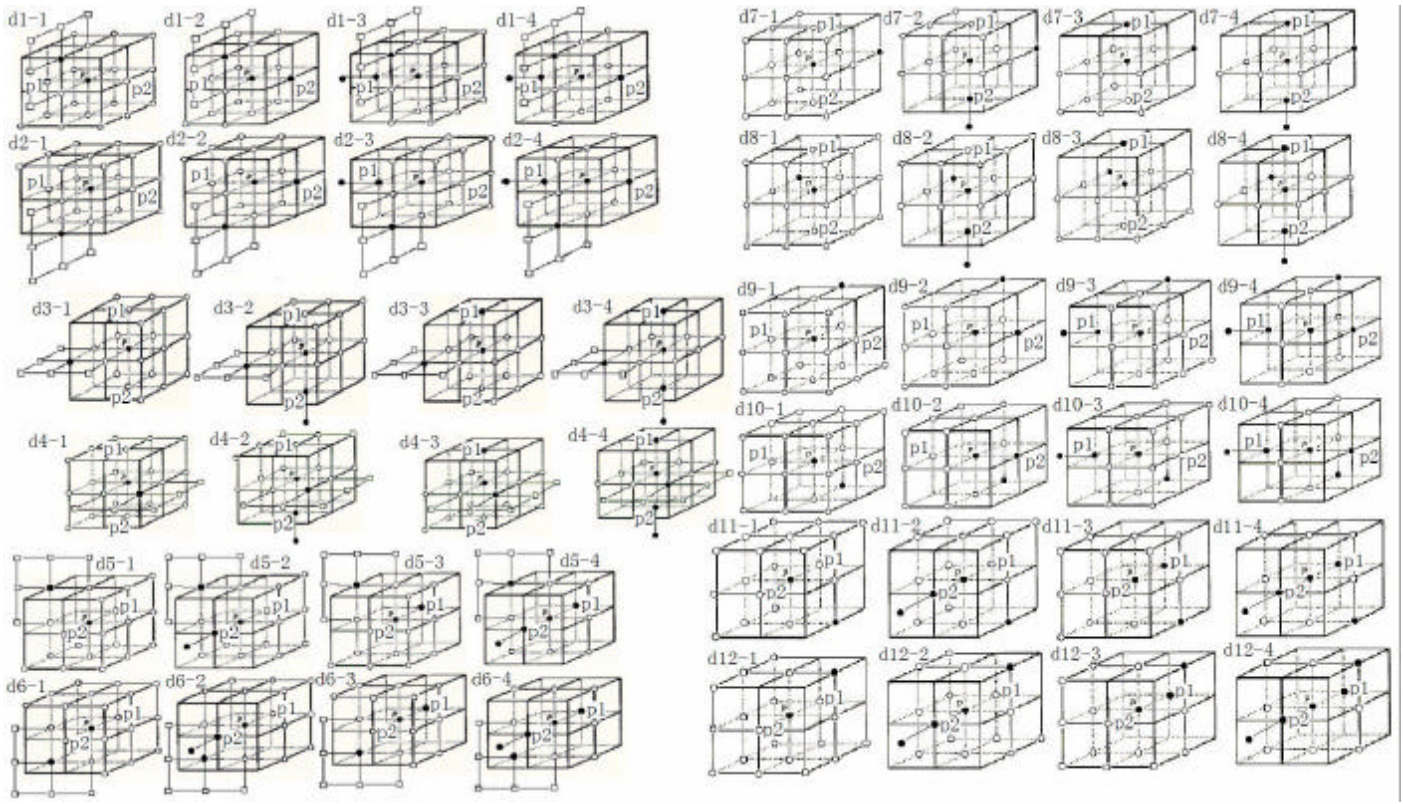

**Figure 10:** The modified deleting templates in Class D. Each template in Class D is changed to three templates, in which (*p1, p2*) are (0, 0), (0, 1) or (1, 0) respectively. At least one point marked ? is an object point.

Figure 11 shows some different results of Ma and Sonka's algorithm and the modified one. Six 26-connected 3D objects are shown in (a). For Ma and Sonka's algorithm, point *c, d* and *e* are deleted, thus the connected object is disconnected after thinning, as shown in (b). For the modified algorithm, point *c* and *e* are deleted, but point *d* will not be deleted, thus connectivity is preserved after the thinning operations, as shown in (c).

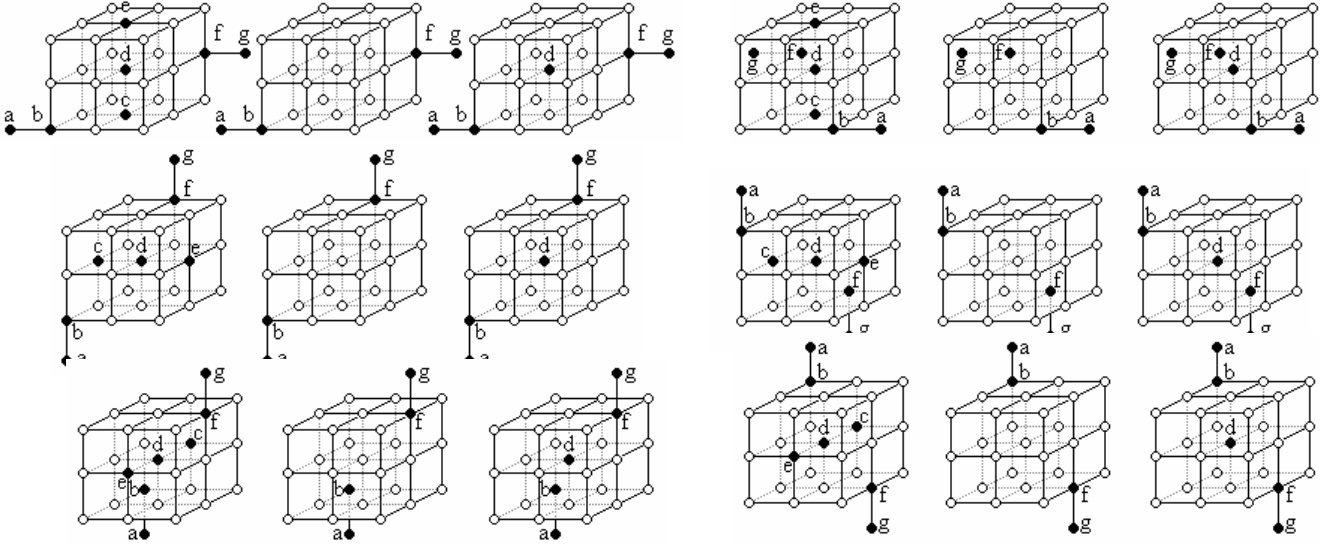

(a)          (b)          (c)              (a)          (b)          (c)

**Figure 11:** A "•" is an object point. A "∘" is a background point. All other points in 3D space are background points. (a) The original 3D object a-b-c-d-e-f-g. (b) The thinning result of Ma and Sonka's algorithm. Point *c*, *d* and *e* are deleted by some templates in Class A and Class D. Thus, the object gets disconnected. (c) The thinning result of the modified algorithm. Points *c* and *e* are deleted by some templates in Class A, but point *d* is not deleted, thus the object is still connected.

Figure 12 shows a different result of Ma and Sonka's algorithm and the modified one. Two "0"s connected by a structure (as shown in Figure 7) are shown in (a). For Ma and Sonka's algorithm, the connected object is disconnected after thinning, as shown in (b). For the modified algorithm, the connected object is still connected after the thinning operations, as shown in (c).

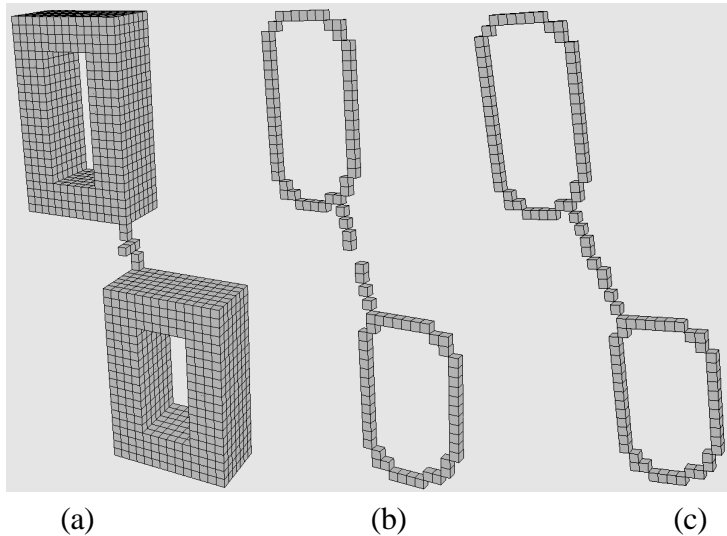

(a)                    (b)                    (c)

**Figure 12:** (a) Original 3D object; (b) Result of Ma and Sonka's algorithm; (c) Result of modified algorithm.

**Verification**

The verification procedure is same as Ma and Sonka's algorithm. LEMMA 3.5 satisfies, therefore, the modified algorithm is connectivity preserving.

## 5. Conclusions

In this paper, a correction of Ma and Sonka's fully parallel 3D thinning algorithm [9] is proposed. The motivation behind this paper is that we found Ma and Sonka's algorithm failed to preserve connectivity for some configurations, which is very important for 3D thinning algorithms. We then studied why Ma and Sonka's algorithm failed and proposed a solution to the problem. Experimental results demonstrate the validity of our solution.

In future work, we will apply our algorithm in a variety of CT and MRI data processing applications.

**Acknowledgements**

The authors gratefully thank Kalman Palagyi, who introduced us to this problem. This work is inspired by our personal communications with him.

**References**

[1] P. J. Besl, R. C. Jain, Three-dimensional object recognition, ACM Computing Surveys, 17 (1), 1985, pp. 75-145.

[2] T. Funkhouser, P. Min, M. Kazhdan, J. Chen, A. Halderman, D. Dobkin, D. Jacobs, A Search Engine for 3D Models, ACM Trans. on Graphics, 22(1), 2003, pp. 83-105.

[3] H. Blum, A transformation for extracting new descriptors of shape, Models for the Perception of Speech and Visual Form, MIT Press, Cambridge, MA, USA, 1967, pp. 362–380.

[4] K. Palagyi, A. Kuba, A 3D 6-subiteration thinning algorithm for extracting medial lines, Pattern Recognition Letters, 19 (7), May 1998, pp. 613-627.

[5] C. Lohou, G. Bertrand, A 3D 12-subiteration thinning algorithm based on P-simple points, Discrete Applied Mathematics, 139 (1-3), April 2004, pp. 171-195.

[6] K. Palagyi, A. Kuba, A parallel 3D 12-subiteration thinning algorithm, Graphical Models and Image Processing, 61 (4), July 1999, pp. 199-221.

[7] P. K. Saha, B. B. Chaudhury, D. D. Majumder, A new shape-preserving parallel thinning algorithm for 3D digital images, Pattern Recognition, 30 (12), December 1997, pp. 1939–1955.

[8] G. Bertrand, Z. Aktouf, A 3D thinning algorithm using subfields, SPIE Proc. of Conference on Vision Geometry, San Diego, CA, USA, 1994, pp. 113–124.

[9] C. M. Ma, M. Sonka, A fully parallel 3D thinning algorithm and its applications, Computer Vision and Image Understanding, 64 (3), November 1996, pp. 420-433.

[10] T. Y. Kong, On topology preservation in 2-D and 3-D thinning, International Journal on Pattern Recognition and Artificial Intelligence, 9 (5), 1995, pp. 813 – 844.

[11] G. Bertrand, Simple points, topological numbers and geodesic neighborhoods in cubic grids, Pattern Recognition Letters, 15(10), 1994, pp. 1003-1011.

[12] P. K. Saha and B. B. Chaudhuri, Detection of 3-D Simple Points for Topology Preserving Transformations with

Application to Thinning, IEEE Trans. Pattern Anal. Mach. Intell., 16(10): 1028-1032, 1994.

[13] T.Y. Kong, On the problem of determining whether a parallel reduction operator for n-dimensional binary images always preserves topology, SPIE Vision Geometry II, Vol. 2060, pp. 69-77, 1993.

[14] C. M. Ma, On topology preservation in 3D thinning, CVGIP: Image Understanding, 59(3), 1994, pp. 328–339.

[15] G. Bertrand. On P-simple points. *Comptes Rendus de l'Acadé mie des Sciences, Série Math.*, I(321):1077-1084, 1995.

[16] A. Chaturvedi, Z. Lee, Three-dimensional segmentation and skeletonization to build an airway tree data structure for small animals, 50 (7), April 2005, Physics in Medicine and Biology, pp. 1405-1419.

[17] M.S. Talukdar, O. Torsaeter, M.A. Ioannidis, J.J. Howard, Stochastic reconstruction, 3D characterization and network modeling of chalk, Journal of Petroleum Science and Engineering, 35 (1-2), July 2002, pp. 1-21.

[18] C. Lohou, Contribution à l' analyse topologique des images (Ph.D. thesis), UNIVERSITÉ DE MARNE-LA-VALLÉE, 2001.

**Errata**

Some templates dX-2 (X=1...12) should be modified.

d1-2 (no change)

d2-2 (no change)

d3-2 (add one more plane at down side, the following is the added plane)

d d d **(d: don't care point, 1: object point, 0: background point)**

d 1 d

d d d

d4-2 (no change)

d5-2 (add one more plane at south side, the following is the added plane)

d d d

d 1 d

d d d

d6-2 (add one more plane at south side, the following is the added plane)

d d d

d 1 d

d d d

d7-2 (add one more plane at down side, the following is the modified template)

0 d d

0 0 d

0 0 0

0 0 1

0 1 0

0 0 0

d d d

d 1 d

d d d

d d d

d 1 d

d d d

d8-2 (add one more plane at down side, the following is the added plane)

d d d

d 1 d

d d d

d9-2 (no change)

d10-2 (no change)

d11-2 (add one more plane at south side, the following is the added plane)

d d d

d 1 d

d d d

d12-2 (add one more plane at south side, the following is the added plane)

d d d

d 1 d

d d d